\documentclass{elsarticle}

\usepackage{amssymb}
\usepackage{amsmath}

\journal{Computers in Human Behavior: Artificial Humans}

\begin{document}

\begin{frontmatter}


\author[1]{Shravan Murlidaran\corref{cor1}\fnref{equal}}
\ead{smurlidaran@ucsb.edu}

\author[1,2,3]{Miguel P. Eckstein\fnref{equal}}
\ead{eckstein@psych.ucsb.edu}

\cortext[cor1]{Corresponding author.}
\fntext[equal]{These authors contributed equally to this work.}

\affiliation[1]{organization={Psychological \& Brain Sciences, University of California, Santa Barbara},
             addressline={},
             city={Santa Barbara},
             postcode={93106},
             state={California},
             country={USA}}

\affiliation[2]{organization={Department of Computer Science, University of California, Santa Barbara},
             addressline={},
             city={Santa Barbara},
             postcode={93106},
             state={California},
             country={USA}}

\affiliation[3]{organization={Department of Electrical and Computer Engineering, University of California, Santa Barbara},
             addressline={},
             city={Santa Barbara},
             postcode={93106},
             state={California},
             country={USA}}

\title{Evolution of  Accuracy and Visual-Cognitive Errors in a Decade of Vision-Language AI Models}

\begin{abstract}
Vision–language models (VLMs) have made remarkable progress in visual reasoning during the last decade. Most evaluations have used simple scenes (MS-COCO) that do not showcase complex human interactions or behaviors,  only a handful of non-curated human descriptions as a benchmark, and have not focused on understanding the model's error types. Here, we introduce the Complex Social Behavior (CSB) dataset, containing 100 images depicting complex social interactions/behaviors. We analyze the progression of scene descriptions over a decade (2017-2025) of VLMs (four pre-Multimodal Large Language Models, MLLMs, and five MLLMs). We evaluate the accuracy of the models and 20 human descriptions relative to a gold standard on the CSB dataset and on a sample from MS-COCO. We analyzed five visual-cognitive error types: object detection, recognition, hallucination, scene understanding, and spatial dependence.  The CSB dataset showed a more pronounced improvement than MS-COCO in scene description accuracy, with pre-MLLMs achieving much lower accuracy than the bottom-ranked human descriptions and MLLMs attaining accuracies similar to the top-ranked human descriptions. We show that MLLMs have eliminated the gap in scene description accuracy between simpler MS-COCO scenes and scenes depicting complex behaviors (CSB). MLLMs have almost eliminated all error types in our tested datasets, except for occasionally relying on different image regions for scene descriptions than humans do (spatial dependence error). We also show that detection, recognition, and hallucination errors have the highest impact on scene description accuracy.  Together, our findings provide a more thorough evaluation of how visual language models have advanced over the last decade. 
\end{abstract}


\end{frontmatter}



\section{Introduction}
\label{sec1}

\begin{figure}[!hbt]
    \centering
    \includegraphics[width=\textwidth]{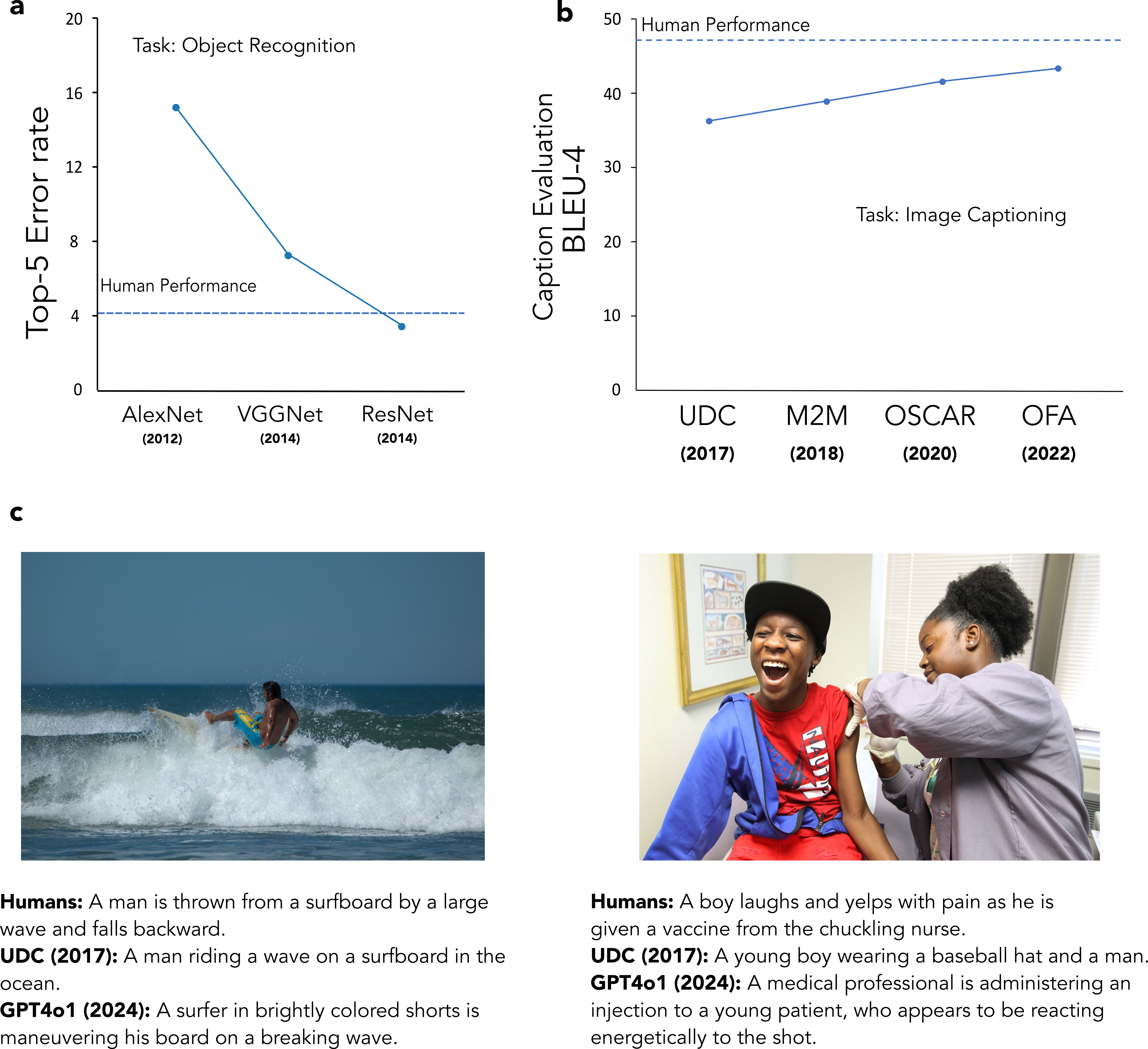}
    \caption{(a) The reported performance of DNNs over the years in object recognition compared to human performance. The performance of the then state-of-the-art models exceeded human performance. (b) Reported performance of Multi-Modal Large-Language Models (MLLMs) in the scene description task on the MS-COCO dataset. The model's performance is comparable to human performance. (c) Model descriptions for scenes with and without social interaction. The image on the left depicts a simple scene in which the models' descriptions are similar to those observed in humans. The image on the right depicts a scene of complex social interactions among humans. We can clearly see that the human description captures the social interaction in the scene. In contrast, the pre-MLLM's description does not capture the social interactions, while MLLMs do. }
    \label{fig:perOfDNNs}
\end{figure}
Artificial Neural Networks have advanced significantly in visual reasoning since the development of deep Convolutional Neural Networks (CNNs) over a decade ago \citep{krizhevsky2012imagenet}. Figure \ref{fig:perOfDNNs}a shows a glimpse of the early years of deep CNNs, with models that could outperform humans in tasks like object recognition. Deep Recurrent Neural Networks \cite{salehinejad2017recent}, which process and generate sequential data like text, combined with deep CNNs and image-text datasets \citep{lin_microsoft_2015, sharma_conceptual_2018, agrawal_vqa_2016, wu2017visual, zellers_recognition_2019} led to the rise of task-specific multi-modal models that can either provide captions \cite{chen2021survey}, answer questions \cite{manmadhan2020visual}, visually reason \cite{he2021interpretable, zakari2022vqa} based on images or generate images based on text \cite{elasri2022image}. The development of transformer architectures \citep{vaswani_attention_2017} and improvement in pre-training methods \cite{han2021pre} gave way to task-agnostic multi-modal models that can perform all vision language tasks with minimal fine-tuning required \cite{du2022survey, zhang2024vision}. In addition to these developments, the growth of data and computational power has accelerated the development of Multi-Modal Large Language Models (MLLMs), which are claimed to achieve human-level reasoning capabilities.
\\\\
Often, images in standard datasets are simple scenes (e.g., an image of a kitchen, desk, or beach scene) that do not include complex human actions or social interactions. Thus, the datasets may not capture the human behaviors and interactions that are critical to everyday visual inferences. Describing these complex scenes accurately might be more challenging for models than describing simple scenes from typical datasets. Figure \ref{fig:perOfDNNs}b shows that the reported performance of the then state-of-the-art vision-language models (based on values reported in the corresponding model papers \cite{anderson2018bottom, cornia2020meshed, li_oscar_2020, wang2022ofa} published between 2017 and 2022) using the BLEU-4 \cite{papineni2002bleu} metric approaches human performance in such datasets. However, Figure \ref{fig:perOfDNNs}c highlights the inability of the 2017 model to capture complex social interactions. The figure showcases depictions provided by a 2017 state-of-the-art image captioning model (UDC \citep{anderson2018bottom}) and a current MLLM (GPT-4o1 \cite{jaech2024openai}) for two scenes: one from the traditional MS-COCO \cite{lin_microsoft_2015} dataset (left) and another from a movie scene portraying complex social interactions (right). Notably, the 2017 model performs reasonably well on the MS-COCO dataset, comparable to current MLLMs, but it fails to capture the complex social interactions in the CSB dataset. This indicates that evaluating the model's progress in scene understanding requires the use of more challenging scene datasets that depict complex human behaviors and interactions.
\\\\
In addition, although vision-language models have improved in their ability to describe scenes, there has been limited work examining which types of errors have decreased over time. Recent research has evaluated properties of these models, such as robustness, bias, fairness, and toxicity \cite{liang2022holistic}. Other studies have analyzed how factors like training data, model architecture, and training or inference procedures influence model decision-making \cite{dang2024explainable}, have examined the shortcomings of VLMs in describing scenes using controlled image perturbations, revealing errors such as misclassification, omission, and incorrect quantity when new objects are inserted into an image \cite{yu2022automated}, and identify factual errors in chart descriptions \cite{huang2023lvlms}.  In this study, we introduce a new taxonomy of visual–cognitive errors to improve understanding of where models struggle with visual reasoning and to assess which errors have been overcome and which persist. The categories include errors in object detection, object recognition \cite{saleh2021occlusion}, hallucinations \cite{rohrbach_object_2019}, scene understanding, and spatial dependence \cite{craig2016review}.
\\\\
Building on this, we have developed the Complex Social Behavior (CSB) dataset, which comprises scenes featuring intricate social interactions and behaviors. Using this dataset, we assess how vision–language models have evolved over the past decade. We also classify the types of errors and examine how they have evolved over the past decade.
\\\\
We compare the performance of nine vision language models developed in the last decade (2017-2025) relative to humans: Up-Down Captioner (UDC) \citep{anderson2018bottom}, Meshed Memory Transformer(M2M) \citep{cornia2020meshed}, OSCAR (\cite{li_oscar_2020}), OFA \cite{wang2022ofa} are four pre-MLLMs, while GPT-4 \cite{achiam2023gpt}, Gemini \cite{team2024gemini}, GPT4-o1 \cite{jaech2024openai}, GPT-5.1, and Gemini3 \cite{team2025gemma} are five MLLMs. Some early models are no longer functional on current platforms, making our evaluation, which is the result of cumulative work beginning in 2019, unique in documenting progress in the evolution of MLLMs. 

As a comparison with our CSB dataset, we conducted the same performance evaluation and error analysis on a separate set of images from the widely used MS-COCO dataset \citep{lin_microsoft_2015}. We evaluate the image descriptions of models and humans against a gold-standard description. Historically, for vision–language models, studies have used automatic metrics to evaluate model descriptions that use sentence structures \cite{papineni2002bleu, kilickaya2016re, luo2022thorough} (BLEU, CIDER, ROUGE). More recently, improved metrics using language embeddings, such as BERTScore \cite{zhang2019bertscore}, have been developed to compare model and human descriptions. Here, we use Gemini/GPT Similarity Index (Gemini-SI/GPT-SI) based on the cosine similarity of description embeddings (similar to BERTScore), which we show correlates better with human ratings than BERTScore (Supplementary Figure \ref{fig:ExtData}).


\section{Methods}\label{sec11}
\begin{figure}[!hbt]
    \centering
    \includegraphics[width=\textwidth]{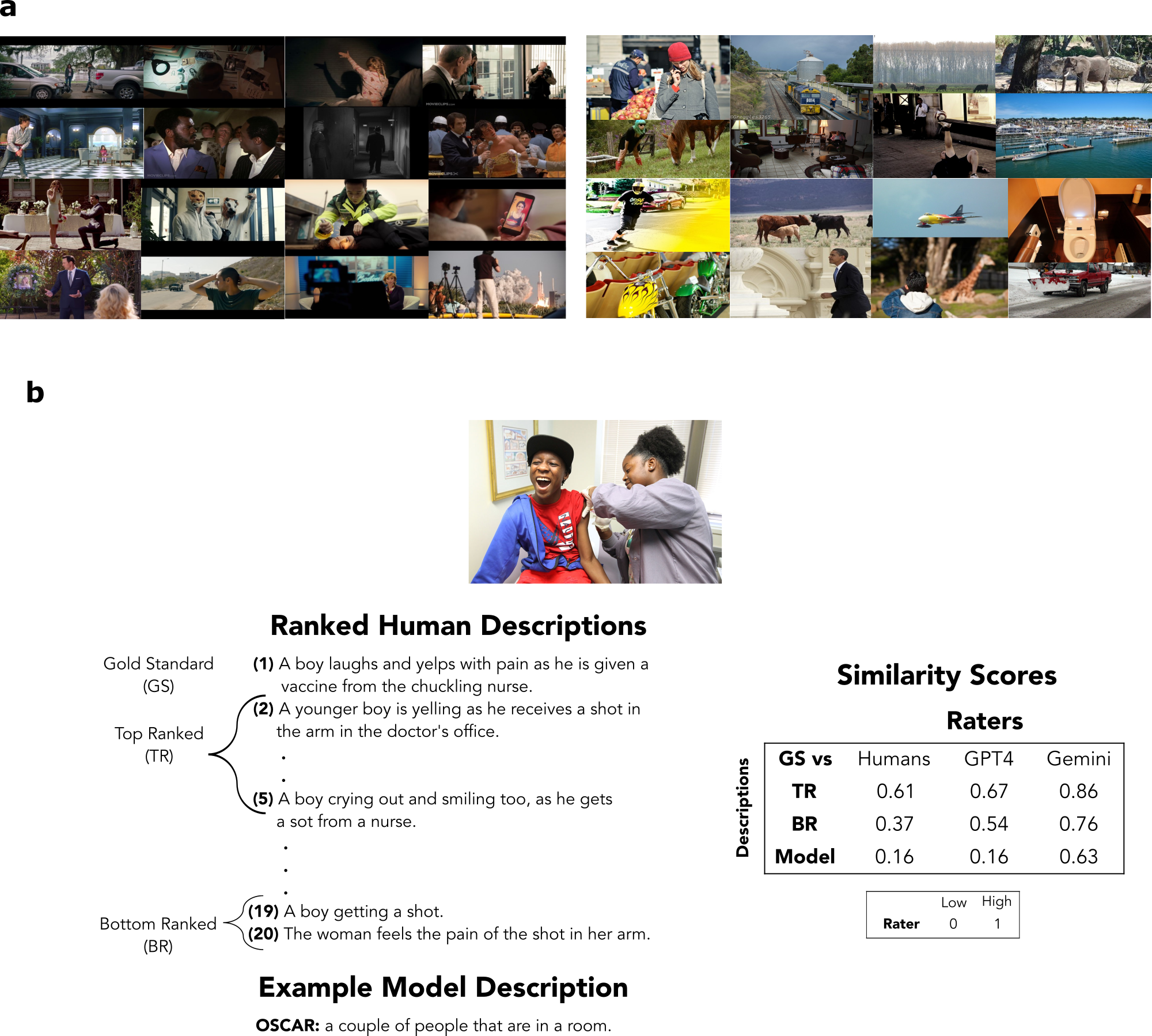}
    \caption{(a)Examples of images from the CSB (top left) and MS-COCO (top right) dataset. The CSB dataset includes images that showcase complex human behaviors and interactions. In contrast, the images from MS-COCO datasets have images with no people or people doing simple activities like walking, sitting, talking, etc. (b) 20 human descriptions are ranked based on how well they describe the image, and the top-ranked description is chosen as ground truth that is used for comparing the similarities of the models and other human descriptions. CSB images shown in this and subsequent figures were sampled from commercially released films and are shown under fair use for non-commercial academic research. Copyright remains with the respective rights holders.}
    \label{fig:ImageExamplesAndFlowchart}
\end{figure}
\subsection{Images}
We used 200 images in this study. Of these, 100 were randomly selected from the MS-COCO dataset. The other 100 images were collected by ten research assistants (RAs). The RAs were instructed to select frames from movie scenes in which they could formulate questions that require reasoning about social interactions or complex human behaviors. Figure \ref{fig:ImageExamplesAndFlowchart}a shows examples of images in the CSB and MS-COCO datasets used in our study. The small sample of images reflects the time-consuming labor required to analyze each model's error types (see section \ref{sec:CME}). We use resampling bootstrap statistics to quantify the variability of model accuracy due to the image sample size.

\subsection{Models}
We used nine different models. Five of them utilize state-of-the-art MLLM architectures (GPT4 \cite{achiam2023gpt}, Gemini \cite{team2024gemini}, GPT4-o1 \cite{jaech2024openai}, GPT-5.1, and Gemini3 \cite{team2025gemma}). These models employ transformer architectures and are trained on vast datasets of image-text pairs, incorporating human feedback through reinforcement learning techniques \cite{chaudhari2024rlhf}. However, limited details about their internal architecture are publicly available.
\\\\
The remaining four models are pre-MLLMs. Among them, three (M2M \cite{cornia2020meshed}, OSCAR \cite{li_oscar_2020}, and OFA \cite{wang2022ofa}) use a ResNet-101 CNN \cite{he2015deep} to extract image features. These features are processed by a Transformer encoder, which represents a joint vision-language space, followed by a Transformer decoder that generates text. The UDC \cite{anderson2018bottom} model also consists of a ResNet-101 CNN image backbone but uses an LSTM (a type of recurrent neural network \cite{yu2019review}) for both encoding and decoding, rather than Transformers. The UDC model is explicitly trained for describing scenes and cannot perform other tasks.
\\\\
M2M, OSCAR, and OFA are task-specific models, but their encoders are pre-trained on image-text pairs, enabling easy fine-tuning across tasks. Additionally, the OFA model includes language input, with text features processed by a language network (BERT). This allows for minimal fine-tuning across different tasks, as the text input provides additional context about the task at hand.

\subsection{Human descriptions of scenes}
\subsubsection{Collection}
We used Amazon Mechanical Turk to collect 20 descriptions for each image. Participants were instructed to describe the scene’s semantic content. Prior to beginning their scene description task, we provided example images and descriptions to ensure that participants understood the expected format.

\subsubsection{Procedure to establish gold standard description}
These descriptions were ranked by a separate group of 50 Mechanical Turk workers. In each trial, participants were shown one image along with five of its 20 descriptions. To ensure attentiveness, two descriptions from other images were randomly included as distractors. Participants were asked to select the description that best matched the image's semantic content.
\\\\
Each participant completed 800 trials (200 images × four description splits). This meant that each participant selected four descriptions per image. To prevent bias from description splits, each participant received a unique split of the 20 descriptions.
\\\\
Descriptions were ranked by the number of votes received as the "best" description. The most frequently chosen description for each image was designated as the gold standard. To reduce comparisons while capturing the variability in human descriptions, we selected the top four (top-ranked) and bottom four (bottom-ranked) descriptions from the remaining 19 descriptions. These were used to compare models and human-generated descriptions against the gold standard description. The similarity comparisons were conducted using LLM-based metrics, as discussed in the results section below. 

\subsection{Input text for Multi-Modal models}
Six (one Pre-MLLM (OFA) and all MLLMs) of the nine models require input text and an associated image, on which the model generates its output. OFA was finetuned to perform scene description using the input text "Describe Image". For the MLLMs, the following input text was used to prompt the models to generate descriptions of the given image.

Input Text Prompt: "Make your best guess of what might be happening in this scene in one sentence. Avoid mentioning objects that do not aid in understanding the context of the scene."

\subsubsection{Gemini-SI/GPT-SI metrics to evaluate description similarity to gold standard description}
Gemini-SI and GPT-SI are LLM-based similarity (LLM-SI) metrics that use language embeddings (3,072 vector size) produced by MLLMs such as Gemini and GPT \cite{achiam2023gpt,team2024gemini} to represent the semantic content of sentences, and compute cosine similarity between these embeddings to quantify semantic similarity. 

\subsubsection{Comparing Gemini-SI/GPT-SI metrics with human raters of scene descriptions}
To evaluate LLM-based similarity (LLM-SI) metrics, we collected human ratings to compare model-generated descriptions (UDC, M2M, and OSCAR) against the gold standard.
\\\\
16 Mechanical Turk workers participated in this experiment. In each trial, participants rated the similarity between three model-generated descriptions and the gold standard. These human ratings were then correlated with the LLM-SI (Gemini-SI/GPT-SI) metrics and BERTScore \cite{zhang2019bertscore}. Supplementary Figure \ref{fig:ExtData}a presents the results of these comparisons. Figure \ref{fig:ImageExamplesAndFlowchart}b depicts the flowchart explaining the description ranking and rating procedure.

\subsection{Classification of model errors}
\label{sec:CME}
\subsubsection{Pre-MLLMs}
Since all pre-MLLMs rely on a ResNet-101 architecture to extract image features, detection and recognition errors were determined based on the ResNet-101 model's bounding box predictions and object classifications.
\\\\
\textbf{Detection Errors: }Objects of relevance that are not mentioned in the description and lack an associated bounding box are classified as detection errors.
\\\\
\textbf{Recognition Errors: }Objects of relevance that are misclassified, or irrelevant objects that are misidentified and included in the description, are classified as recognition errors.
\\\\
\textbf{Scene Understanding Errors: }Objects of relevance that are correctly recognized but omitted from the description are considered scene understanding errors. 
\\\\
\textbf{Hallucinations: } Objects that are not present in the scene or wrongly classified but are mentioned in the description of the model constitute hallucinations \cite{rohrbach_object_2019}.

\subsubsection{MLLMs}
Since the architectures of MLLMs are not publicly disclosed, we identified images where model-generated descriptions deviated from human descriptions and then posed targeted questions to categorize the error types.
\\\\
\textbf{Detection Errors:} If an object of relevance is missing from the description, we referenced nearby landmarks and asked whether the model could see an object near a specific landmark. If the model failed to detect it, this was categorized as a detection error. Example question: "Is there an object in the hands of the person standing next to a wall?"
\\\\
\textbf{Recognition Errors:} We asked the model to name the missed object. If it misclassified the object, it was marked as a recognition error. Example question: "What is the name of the object in the hands of the person standing next to a wall?"
\\\\
\textbf{Scene Understanding Errors:} If the model correctly recognized an object but still omitted it from its description, it was categorized as a scene understanding error.
\\\\
\textbf{Hallucinations:} If the model described an object that was not present in the scene and was not simply a misclassification of another object, it was considered a hallucination. To verify whether an object was wrongly recognized, we analyzed the model's description, identified potential locations based on landmarks, and asked questions similar to those used for detection and recognition errors.

\subsubsection{Spatial dependence errors using masking paradigm:}
\begin{figure}[!hbt]
    \centering
    \includegraphics[width=\textwidth]{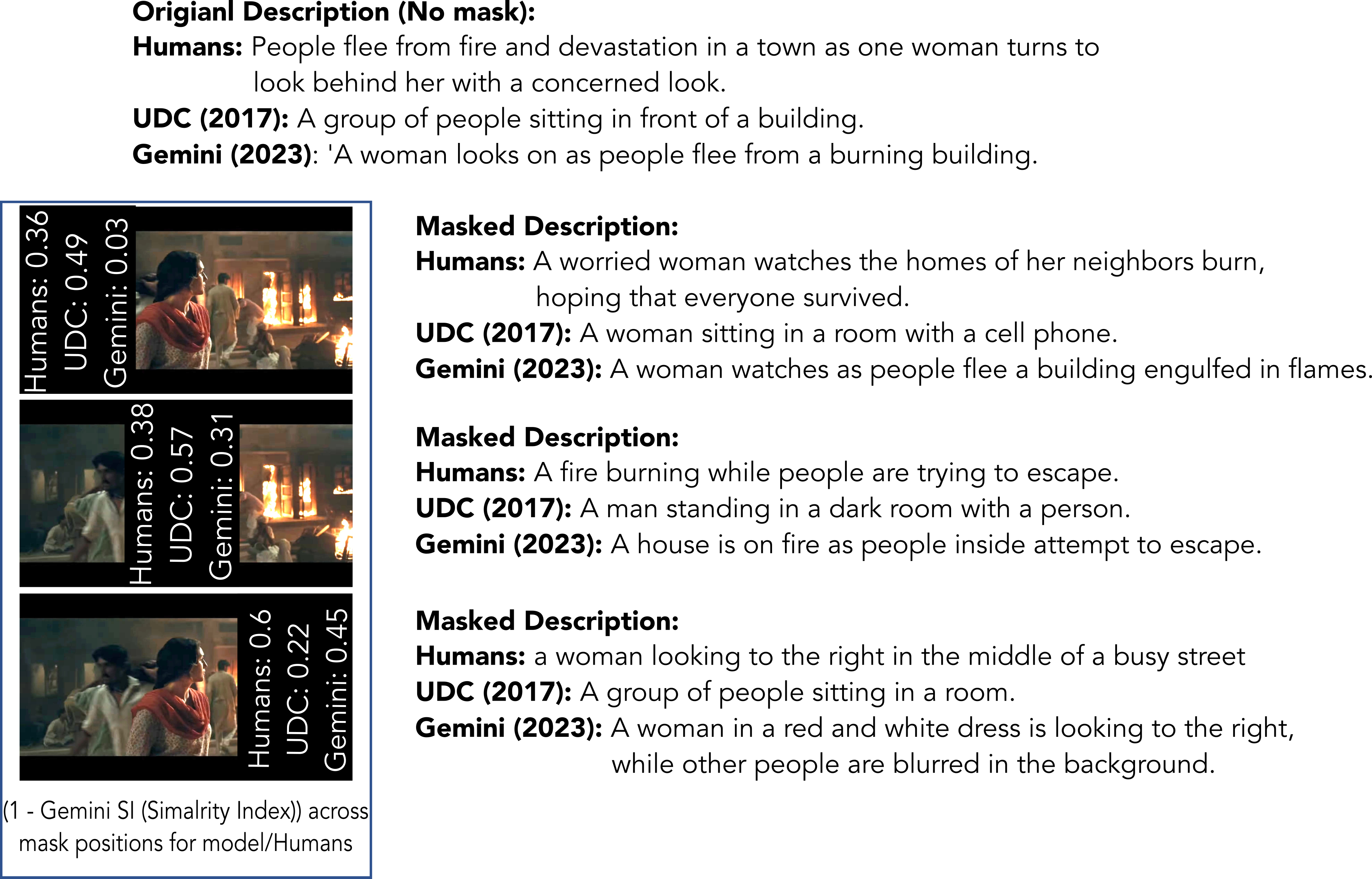}
    \caption{Example of the masking procedure used in our study to quantify spatial dependence errors. The masking procedure identifies the importance of a region of an image to the scene description by evaluating the impact on the scene description when that region is masked(occluded). This procedure is done for both human and model descriptions. The Spatial Dependence Error (SDE) uses the masking procedure to identify whether the model and the humans match (correlation) in the regions that are most important for them to provide a description.}
    \label{fig:SDE_flowchart}
\end{figure}
The last error type measured the degree to which models and humans agreed on which image regions contributed to the description. If the model and humans relied on different parts of the images for the scene description, we classified this as a spatial dependence error.
\\\\
To assess the dependence of the scene description on different image regions, we used a masking paradigm.
Occlusion masks consisted of a black opaque spatial window with a soft edge, covering approximately one-third of each image. These masks varied in position. We created a total of 600 masked images (200 images × 3 masking positions).
\\\\
Descriptions for these masked images were collected from 60 Mechanical Turk workers (20 per masking condition). Each worker saw only one masked version of each image to prevent memory effects from influencing their descriptions.
\\\\
The masked descriptions were compared with the gold standard (unmasked) descriptions to assess which image regions were most important for human perception. For the models, we generated descriptions for each masked version and compared them to descriptions generated for the unmasked images. Figure \ref{fig:SDE_flowchart} illustrates an example of this masking procedure used to analyze which image regions influence the scene descriptions for models and humans. Similarity evaluations were performed using the LLM-based metrics discussed above.
\\\\
If the image regions most critical for descriptions differed significantly between models and humans, we categorized this as a Spatial Dependence Error (SDE). We quantified this by correlating similarity ratings between model-generated and human-generated descriptions for the three masked regions and comparing them to the average inter-human correlation for these regions. If the model-human correlation was two standard deviations below the inter-human correlation, it was classified as a spatial dependence error.

\subsection{Statistical analysis}
Given the small dataset (200 images), we employed bootstrapping to estimate the variability in our measures (scene description similarity to the gold standard and error rates) across images and thus their generalization to the population of images. We used a total of 10,000 image bootstrap resamples per analysis. Since model performance comparisons involved more than 15 statistical comparisons, we corrected significance levels for the False Discovery Rate (FDR) using the Benjamini-Hochberg method \cite{benjamini1995controlling}. The adjusted p-values (q-values) were reported accordingly.

\section{Results}\label{sec2}

We selected a dataset of 100 images depicting complex social interactions and behavior (CSB) and 100 images from the widely used MS-COCO dataset \citep{lin_microsoft_2015}.  
Figure \ref{fig:ImageExamplesAndFlowchart}a presents examples of images used in our study. We utilized nine vision-language models: four pre-MLLMs (UDC, OSCAR, M2M, OFA) and five MLLMs (GPT-4, Gemini, GPT-4o1, GPT5.1, and Gemini3). All pre-MLLMs evaluated contain an object detection component and employ a fully trained, faster R-CNN model. OSCAR, M2M, and OFA use variants of the Transformer architecture to encode a combined vision-language feature space, from which descriptions of a given input image are generated after fine-tuning. UDC, on the other hand, uses LSTMs for this purpose. The Multi-Modal Large Language Models (MLLMs) also rely on Transformer architectures, but their model details are not publicly available.

\subsection{Evolution of vision-language models in providing human-level descriptions for scenes depicting complex social interaction}

\subsubsection{Gold standard descriptions}
To obtain high-quality gold standard descriptions, we collected 20 human-generated descriptions for each image in our dataset. A separate group of 50 participants then ranked these descriptions based on how well they matched with each scene. The top-ranked description for each image was designated as the gold standard and used to compare with model-generated descriptions and other human descriptions. Among the remaining 19 human descriptions, the top four were classified as top-ranked, and the bottom four as bottom-ranked. Figure \ref{fig:ImageExamplesAndFlowchart}b(left) illustrates an example image along with human and model-generated descriptions.

\subsubsection{Metric of semantic similarity for description comparisons}
To compare model performance with human descriptions on the CSB and MS-COCO datasets, we measured the similarity between model-generated descriptions and the gold standard using the cosine similarity of text embeddings. These embeddings were obtained from LLMs such as Gemini (Gemini-Similarity Index, Gemini-SI) \cite{team2024gemini} and GPT (GPT-Similarity Index, GPT-SI) \cite{achiam2023gpt}. Text embeddings are numerical vectors representing the semantic meaning of a sentence in a high-dimensional space. In this space, sentences with similar meanings are represented by vectors that are closer to each other.
\\\\
We validated the use of GPT-SI and Gemini-SI by comparing them with human ratings. To this end, we collected similarity ratings from 20 human raters for scene description pairs comparing three models (UDC, M2M, and OSCAR) to the gold standard. We found that the average correlation of GPT-SI and Gemini-SI with individual human raters was higher than the average correlation among human raters (bootstrap, p $<$ 0.001) for all three models (see Supplementary Figure \ref{fig:ExtData}a). Figure \ref{fig:ImageExamplesAndFlowchart}b(right) illustrates the ratings provided by humans and LLM-SI (Gemini-SI and GPT-SI) for a model-generated description compared to the gold standard.
\begin{figure}[!hbt]
    \centering
    \includegraphics[width=0.8\textwidth, scale = 0.8]{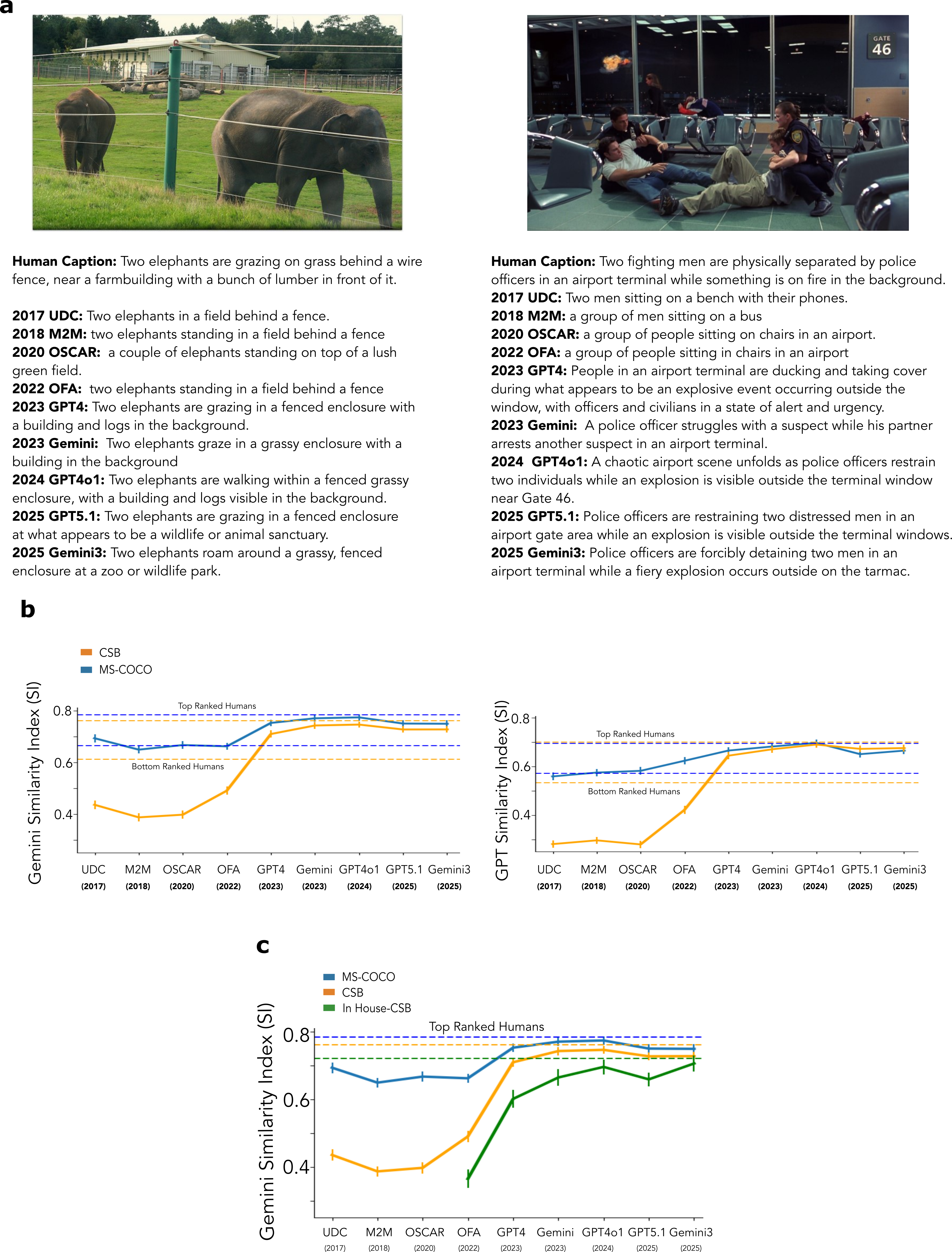}
    \caption{(a) Evolution of model description of a sample image from the MS-COCO dataset(left) and the CSB dataset (right). The model descriptions are similar to human descriptions for the MS-COCO image. In contrast, pre-MLLMs perform poorly on the CSB dataset compared to humans, whereas MLLMs produce descriptions similar to those of humans. (b) The performance of the models and humans was evaluated using Gemini-SI(left) and GPT-SI(right) metrics in both datasets. Both metrics show that the pre-MLLMs had significantly lower performance in the CSB dataset than in the MS-COCO dataset, and these models were significantly worse than the bottom-ranked human descriptions in the CSB dataset, while they were at par with the bottom-ranked human descriptions in the MS-COCO dataset. The MLLMs performed similarly in both datasets and were on par with the top-ranked human descriptions. (c) Performance of MLLMs and one pre-MLLM (OFA) on the in-house complex social behavior dataset in comparison to our results from the dataset used in this study. This dataset was used to ensure the MLLMs' performance is not due to the possibility of seeing these images in our dataset during training. We see that the model's performance is similar to human performance, even on the in-house-CSB dataset.}
    \label{fig:ModelPerformanceEval}
\end{figure}
\subsubsection{Human-model descriptions' semantic similarity, 2017-2025}
Figure \ref{fig:ModelPerformanceEval}a shows example images from the MS-COCO dataset and the CSB dataset, along with the description provided by the models developed in the last ten years. The example illustrates that, even in 2017, image caption models could describe many scenes in the MS-COCO dataset, and that the introduction of MLLMs has improved these descriptions. However, the more challenging CSB dataset shows significant improvements in visual common-sense reasoning capabilities with the introduction of MLLMs.  Figure \ref{fig:ModelPerformanceEval}b shows the model performances as evaluated by Gemini-SI (left) and GPT-SI (right) in both datasets. On average, the pre-MLLMs perform significantly worse in the CSB images than the MS-COCO images (bootstrap, p $<$ 0.0001), while the MLLMs perform similarly in both image sets. All the MLLMs' performances are close to the top-ranked human performances in both datasets (dotted lines). The pre-MLLMs perform at the level of the bottom-ranked humans on the MS-COCO dataset and are significantly lower than the bottom-ranked humans (bootstrap, p $<$ 0.0001) on the CSB dataset.
\\\\
One possible explanation for the MLLMs' high performance is that images in the CSB dataset, which consisted of movie scenes available online, may have been part of the MLLM's vast training data. To discount such an explanation, we also tested all the MLLMs and one pre-MLLM (OFA) on an in-house dataset of 100 images (see Supplementary Figure \ref{fig:ModelPerformanceEval}c) not posted online (as of these models' release), and thus unseen by the models during training. We found similar trends for the in-house images as for the movie scenes, with MLLM performance close to that of humans, while the OFA model performed significantly lower (bootstrap, p $<$ 0.0001) than humans.
\\\\

\subsection{What kind of errors do models make?}

To better understand the evolution of the models, we analyzed the types of errors each model made when generating descriptions. A typical vision-language model consists of two main components: an object detector and an encoder-decoder architecture, both of which can introduce errors in their descriptions. We categorized these errors based on concepts from vision science and cognitive psychology.

\begin{figure}[!hbt]
    \centering
    \includegraphics[width=\textwidth, scale = 0.4]{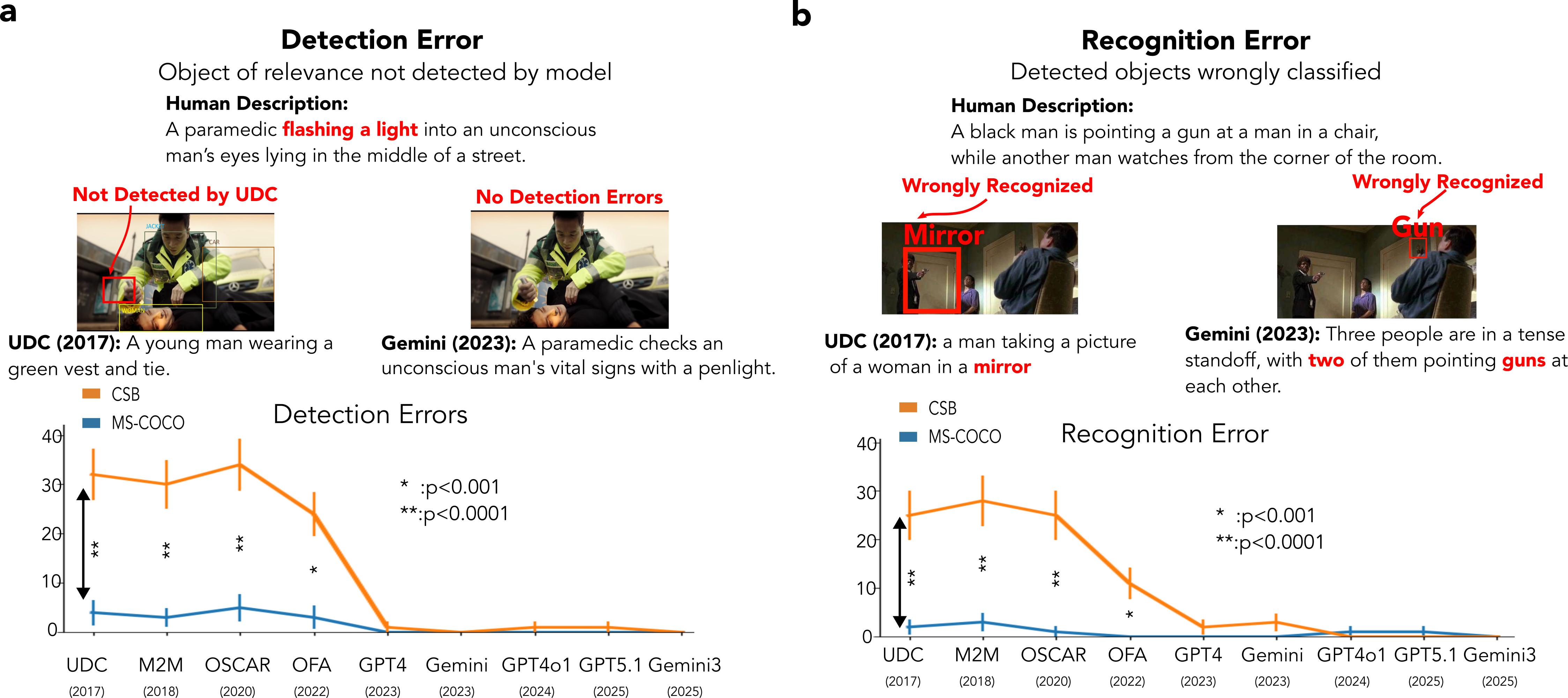}
    \caption{(a) Example of detection errors made by a pre-MLLM and an MLLM. The object of relevance is defined based on its presence in the human gold standard description. The UDC model fails to detect the flashlight, whereas the Gemini model does. Pre-MLLMs make significantly more errors on the CSB dataset than on the MS-COCO dataset, whereas MLLMs are mostly accurate at detecting all objects. (b) Example of recognition errors made by a pre-MLLM and an MLLM. The models incorrectly identify the detected objects. The UDC model incorrectly identifies the doorway as a mirror, while the Gemini model identifies the lamp on the wall as another gun. As with detection errors, pre-MLLMs exhibit significantly higher errors in the CSB dataset than in the MS-COCO dataset.}
    \label{fig:DetRegExamples}
\end{figure}

\subsubsection{Detection errors}
Detection errors occur when a model fails to detect relevant objects in an image. An object is considered relevant if it appears in the gold standard human description and is present in the scene. As illustrated in Figure \ref{fig:DetRegExamples}a, humans identified a flashlight in their descriptions, making it a relevant object. However, the UDC (pre-MLLM) model failed to detect it, whereas the Gemini (MLLM) model correctly identified it. Pre-MLLMs made significantly more detection errors in the CSB dataset compared to the MS-COCO dataset (bootstrap, p $<$ 0.001), whereas MLLMs detected objects accurately in most cases.

\subsubsection{Recognition errors}
Recognition errors occur when a model: 1) incorrectly classifies a relevant object, or (2) misidentifies an irrelevant object and incorrectly mentions it in the description. Figure \ref{fig:DetRegExamples}b provides an example: the UDC (pre-MLLM) model detected a doorway but misidentified it as a mirror, whereas the Gemini model misidentified a lamp as a second gun. Pre-MLLMs exhibited significantly higher recognition errors in the CSB dataset than in the MS-COCO dataset (bootstrap, p $<$ 0.001), with a decreasing trend in errors over time. All models performed well at recognizing objects in the MS-COCO dataset.

\begin{figure}[!hbt]
    \centering
    \includegraphics[width=\textwidth, scale = 0.4]{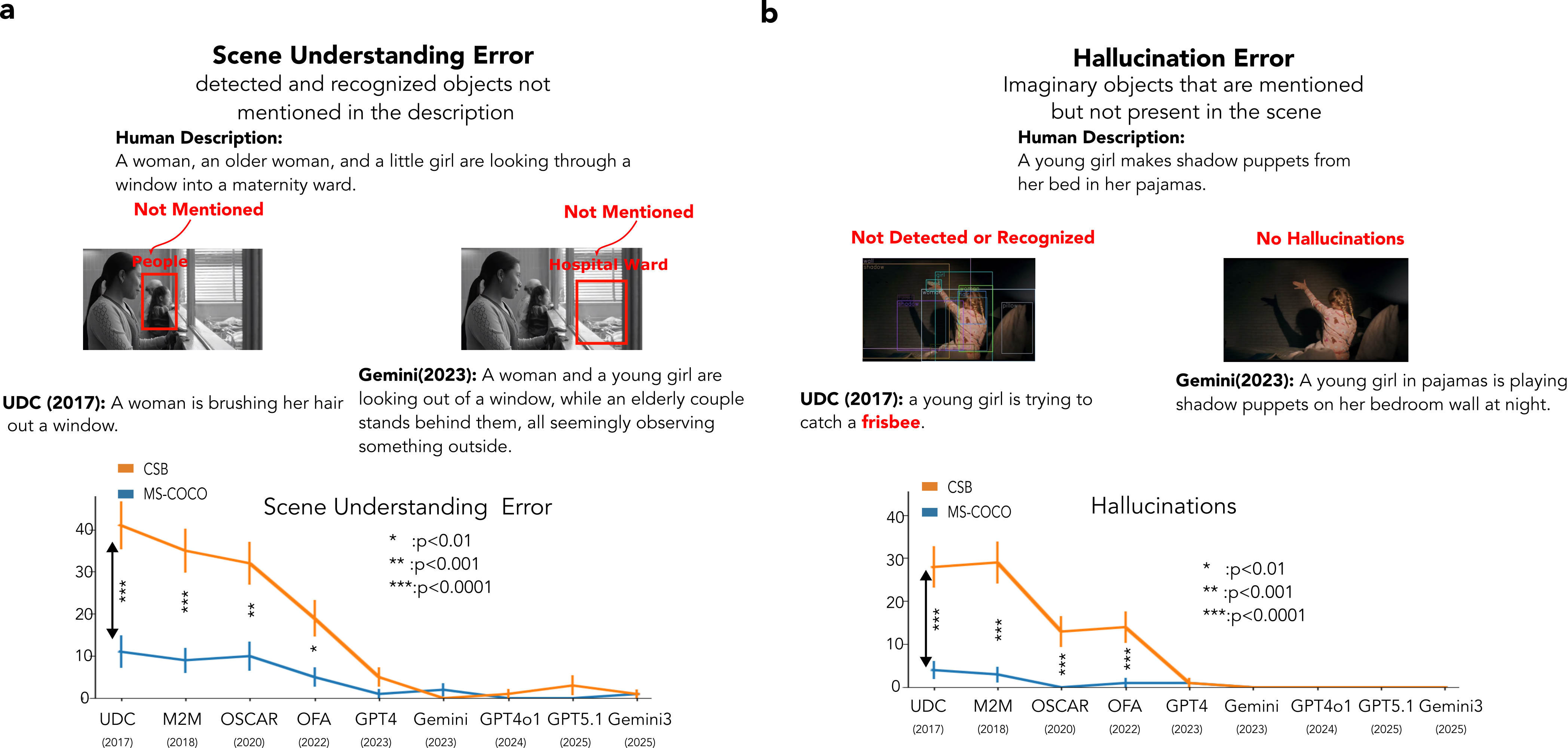}
    \caption{(a) Example of scene understanding errors made by a pre-MLLM and an MLLM. These errors occur when the model fails to mention the correctly recognized relevant objects in its description. The UDC model correctly recognizes the people in the image but fails to mention them in its description. Similarly, the Gemini model correctly identifies that it is a hospital ward but fails to mention it in its description. We observe that the pre-MLLMs make significantly more scene understanding errors in the CSB dataset than in the MS-COCO dataset. In contrast, MLLMs have similar errors in the MS-COCO dataset. (b) Example of hallucinations made by a pre-MLLM and an MLLM. Hallucinations are objects mentioned in the model's description but are neither present in the image nor incorrectly recognized by the model. The UDC model hallucinates a frisbee when it is absent or misrecognized. The Gemini (MLLM) model has no hallucinations in our dataset. The pre-MLLMs have significantly higher errors on the CSB dataset than on the MS-COCO dataset, while the MLLMs exhibit few hallucinations in both datasets.}
    \label{fig:SUHalExamples}
\end{figure}

\subsubsection{Scene understanding errors}
Scene understanding errors occur when a model recognizes relevant objects but fails to mention them in its description. As shown in Figure \ref{fig:SUHalExamples}a, the UDC model recognized people in the scene but did not mention them in its description, while the Gemini model recognized the hospital ward but failed to include it in its description. Pre-MLLMs had significantly higher scene understanding errors in the CSB dataset than in the MS-COCO dataset (bootstrap, p $<$ 0.01), whereas MLLMs exhibited few such errors across both datasets.

\subsubsection{Hallucinations}
Hallucinations \citep{rohrbach_object_2019} occur when a model describes an object that is neither recognized by the model nor present in the image. These objects can be described as 'imaginations' of the model, which may occur when similar images in the model's pretraining dataset contain the hallucinated object.  Figure \ref{fig:SUHalExamples}b shows an example where the UDC model incorrectly mentioned a frisbee. Pre-MLLMs made significantly more hallucination errors in the CSB dataset than in the MS-COCO dataset (bootstrap, p $<$ 0.0001), whereas MLLMs had very few hallucinations.

\begin{figure}[!hbt]
    \centering
    \includegraphics[width=0.84\textwidth, scale = 0.5]{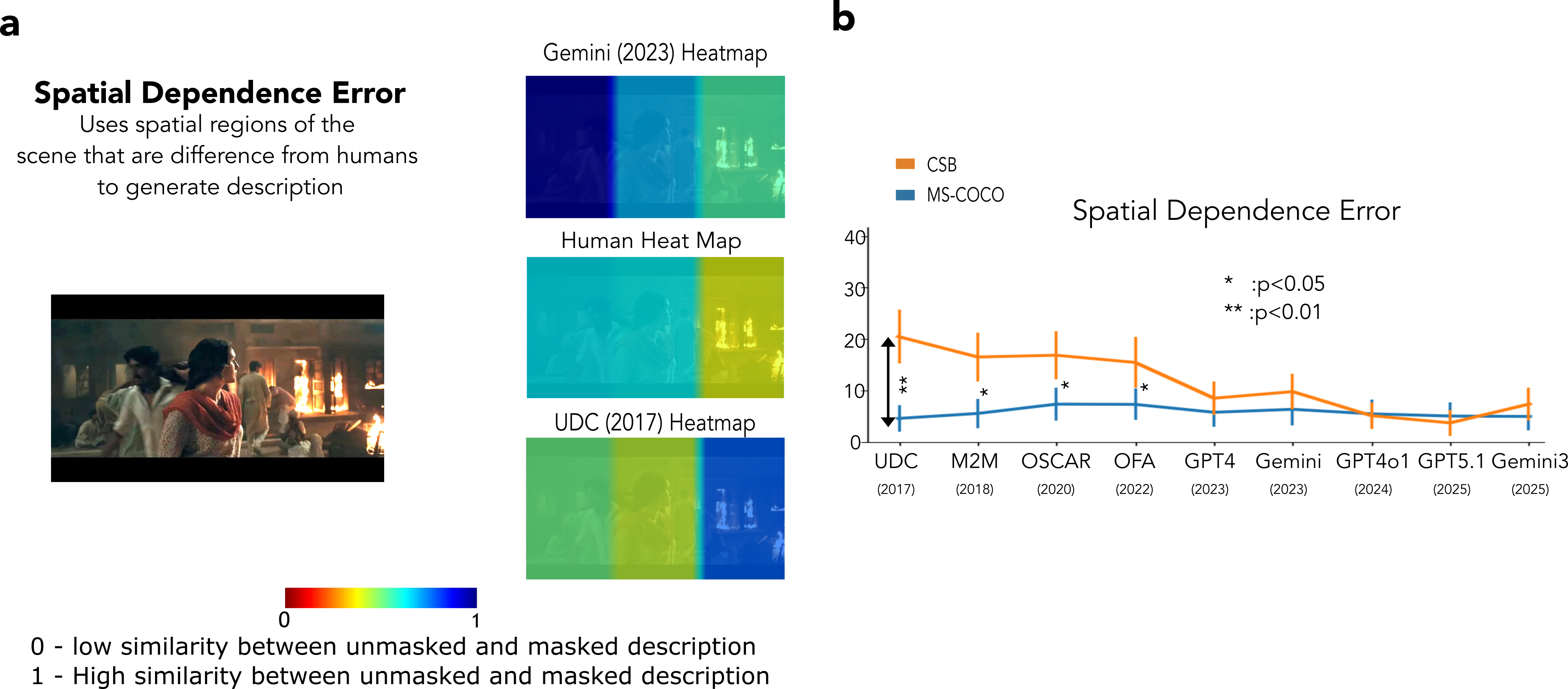}
    \caption{(a) Example heatmaps showcasing the regions that are important for humans, UDC (pre-MLLM) and Gemini (MLLM). The MLLM and humans agree on the regions important for understanding the scene, whereas the pre-MLLM differs, leading to a spatial dependence error. (b) Summarizes the errors made by the models tested on our dataset. The spatial dependence errors (SDE) are significantly high across all pre-MLLMs, and the SDE errors in the CSB dataset are decreasing. The SDE errors are nearly identical across all models on the MS-COCO images.}
    \label{fig:SDE}
\end{figure}

\subsubsection{Spatial dependence errors}
Spatial dependence errors (SDE) occur when humans and models rely on different image regions to generate descriptions. To estimate this, we masked one-third of the image at a time. We collected descriptions from humans and the models after masking each image region and compared the similarity of these descriptions to the corresponding gold-standard descriptions for the models/humans (see methods). Figure \ref{fig:SDE_flowchart} shows a detailed overview of the masking procedure. A lower similarity rating for a masked region indicates that masking that region significantly affects the generated description. When the critical regions used by humans and models do not align, an SDE is recorded. LLM-SIs (GPT-SI and Gemini-SI) are used to assess these similarities. 

In the example shown in Figure \ref{fig:SDE}a, heatmaps illustrate the regions important for description generation. Humans and the UDC (pre-MLLM) model rely on different regions, whereas the Gemini (MLLM) model relies on regions similar to those chosen by humans. Figure \ref{fig:SDE}b demonstrates that SDE errors are considerably higher for all models in the CSB dataset. However, over the years, errors in this dataset show a downward trend across models. In contrast, SDE errors in the MS-COCO dataset remain relatively stable across all models.

\begin{figure}[!hbt]
    \centering
    \includegraphics[width=0.76\textwidth, scale = 0.5]{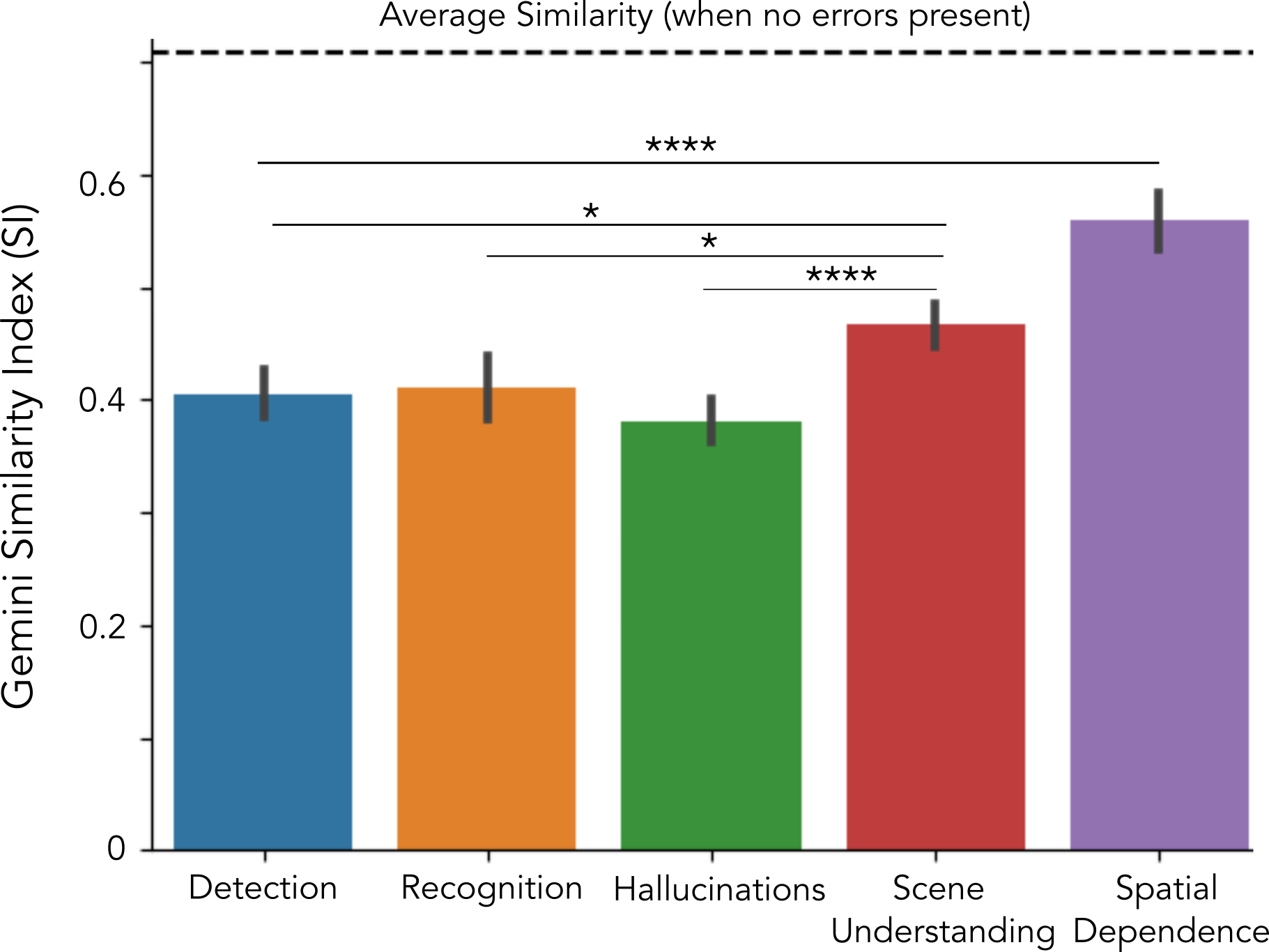}
    \caption{(a) Gemini-SI as a function of the error types averaged across all models and images with the error type present. The dashed line indicates the average similarity across all model-image pairs with no errors in their descriptions. Detection, recognition, and hallucination errors most affect the model's scene description, followed by scene understanding and spatial dependence errors.}
    \label{fig:ErrvsRatings}
\end{figure}

\subsubsection{Impact of model errors on similarity rating}
To quantify how different error types affect the similarity between human and model descriptions, we computed the average Gemini-SI score across all model–image pairs and grouped the results by the error type present. In Figure \ref{fig:ErrvsRatings}, the dashed line indicates the baseline average similarity score of 0.71 for images with no identified errors. Detection, recognition, and hallucination errors produced the largest reductions in similarity, lowering the average Gemini-SI to 0.405, 0.411, and 0.381, respectively. In contrast, scene understanding errors had a significantly smaller impact on description similarity than detection, recognition, and hallucination errors (Gemini-SI = 0.467, bootstrap, p < 0.04), while spatial dependence errors showed the least impact overall (Gemini-SI = 0.560, bootstrap, p < 0.0001).
\\\\
Overall, these findings indicate a significant difference in error rates between the MS-COCO and CSB datasets. In particular, pre-MLLMs on the MS-COCO dataset showed lower detection, recognition, and hallucination error rates than scene understanding error rates. For the CSB dataset, the pre-MLLMs' detection, recognition, and hallucination error rates were closer to the scene understanding error rates and might be related to the critical role of smaller objects in the CSB dataset.  In addition, the MLLMs have reduced rates across the board for all error types for both datasets. The only error type that remains significant for MLLMs is the spatial dependence error, suggesting that MLLMs can sometimes rely on different image regions than humans to describe scenes.

\section{Discussion}\label{sec12}
Comparisons between humans and models span a wide range of tasks, including contour detection, recognition performance under visual distortions \citep{dodge2017study}, visual reasoning with synthetic tasks \citep{firestone_performance_2020}, abstraction and reasoning \citep{chollet_measure_2019}, language representation \cite{jones2024multimodal}, visual search in real-world scenes \citep{eckstein_humans_2017}, aerial scene understanding \citep{deza_assessment_2019}, image captioning \cite{wang2019describing, wang2020towards}, perception of gaze \cite{han2021gaze} and lesion detection in medical images \citep{lago_under-exploration_2021, jonnalagadda2025convolutional}.
\\\\
The present study assessed the evolution of vision-language models over the last decade and compared them with human capabilities for understanding scenes. We selected four pre-Multi-Modal Large Language Models (pre-MLLMs) (UDC \cite{anderson2018bottom}, M2M \cite{cornia2020meshed}, OSCAR \cite{li_oscar_2020}, and OFA \cite{wang2022ofa}) and five MLLMs (GPT4 \cite{achiam2023gpt}, Gemini \cite{team2024gemini}, GPT4-o1 \cite{jaech2024openai}, GPT-5.1, and Gemini3 \cite{team2025gemma}) that cover the range of state-of-the-art models over the last decade. Traditional datasets used in the computer science field \cite{lin_microsoft_2015, sharma_conceptual_2018, agrawal_vqa_2016, wu2017visual, zellers_recognition_2019} often contain images that are simple to understand, and all the models over the last ten years perform relatively well, even though it is evident that the current MLLMs are much superior to the other models. We created the Complex Social Behavior (CSB) dataset, depicting complex human behaviors and social interactions. Additionally, we selected 100 representative images from the MS-COCO dataset \cite{lin_microsoft_2015} and evaluated the models' performance at describing images in both datasets. 
\\\\
Overall, our findings indicate that the assessment of how models have evolved over the last decade varies substantially depending on the image set: MS-COCO versus CSB. Assessment using the CSB images shows the significant improvements between pre-MLLMs and MLLMs for images depicting complex human behavior and social interactions. The assessment with MS-COCO images shows a more modest improvement. Furthermore, our findings highlight that MLLMs have narrowed the performance gap between the MS-COCO and CSB images. Pre-MLLMs exhibit a large performance gap between the two datasets. MLLMs' performance is nearly identical across the simpler MS-COCO and CSB datasets. Comparison with humans on the CSB dataset also shows that the performance of pre-MLLMs is below that of the bottom-ranked humans. In contrast, MLLMs' performance is close to the top-ranked humans. 
\\\\
Aside from performance assessment, it is important to understand across image sets to facilitate future model improvements and help users gain trust in model outputs. To this end, the field of explainable AI (XAI) \cite{xu2019explainable} has introduced many approaches for understanding how models make decisions. Some methods examine how changes in certain features affect the model’s output, while others estimate how much each feature contributes to the final prediction \cite{ribeiro2016should, selvaraju2017grad, cambria2024xai}. There are also approaches that organize examples in a dataset into prototypes, which represent typical cases, and criticisms, which highlight where the model struggles \cite{kim2016examples}. Most of these approaches focus on explaining individual inputs and therefore do not provide an easy way to obtain statistical summaries of explanations across an entire dataset. Moreover, since the architectures of current MLLMs are not publicly available, it is especially important to develop techniques that do not rely on access to internal model components. To understand the shortcomings of these models, we developed a novel technique to classify model errors in image description generation. These errors included object detection errors, object recognition errors \cite{saleh2021occlusion}, hallucinations \cite{rohrbach_object_2019}, scene understanding errors, and spatial dependence errors \cite{craig2016review}. 
\\\\
We found that for images depicting complex social behaviors, there is a significant reduction across all error types, and errors are almost negligible for current MLLMs. MLLMs have reduced errors for these CSB images to comparable levels to the MS-COCO, while pre-MLLMs show much larger error rates for all error types for the CSB images relative to MS-COCO. The one exception is the spatial dependence errors, which have reduced for the CSB dataset but remain a source of error for the MLLMs. This result suggests that humans and MLLMs can sometimes rely on different parts of the image to generate the scene description. 
\\\\
Another result of our analysis is that detection, recognition, and hallucination errors contribute the most to reducing semantic similarity to the gold standard.  The one error type remaining in MLLMs (spatial dependence) contributes the least to reducing semantic similarity between the model description and the gold standard. This persistent spatial dependence error aligns with recent work by \cite{rosenberg2026limits}, who found that affordance-related tasks requiring spatial and agent-centered reasoning remained uniquely resistant to improvement across 18 VLMs, even with prompt engineering and newer model releases. Their findings suggest this reflects a deeper difficulty in reasoning about embodied agents within three-dimensional scenes from two-dimensional images.
\\\\
A limitation of the current study is that we only evaluated 200 images. This might be considered a small sample compared to large dataset evaluations.  Our main bottleneck is the analysis of error types, which requires manually probing the model for each image in the dataset. Detection, recognition, and scene understanding errors are all handled manually for pre-MLLMs. The MLLMs were asked a series of questions to determine the object detection, recognition, vision-language, and hallucination errors. Still, our analysis accounted for variability across images by using a bootstrap resampling method to assess statistical differences between the CSB and MS-COCO datasets and other effects.
\\\\
Our findings should not be interpreted as implying that MLLMs have achieved human-level performance across all perceptual tasks. Others have revealed the remaining limitations of MLLMs. Tong et al. \cite{tong2024eyes} introduced a framework using question answering to analyze the errors made by MLLMs that are induced by their visual components, while Zhang et al. \cite{zhang2025mlms} introduced a new dataset to study shortcomings of MLLMs in reasoning and cognitive understanding of sarcasm in image-associated text, a capability that requires more than understanding the visual details of a scene to achieve good performance. Tong et al. \cite{tong2024eyes} showed that although these models perform well on reasoning tasks, they still lack robust visual grounding and struggle with fundamental perceptual patterns such as orientation, counting, and spatial relations. Similarly, Zhang et al. \cite{zhang2025mlms} demonstrated that most state-of-the-art MLLMs exhibit a significant gap in sarcasm understanding, suggesting that visual detection of elements is not equivalent to understanding the scene. Our technique, like those of Tong et al. \cite{tong2024eyes} and Zhang et al. \cite{zhang2025mlms}, is model-agnostic and operates solely on model inputs and outputs, enabling us to analyze how shortcomings in different model components contribute to errors in generated descriptions.  
\\\\
Overall, we have developed a novel technique to assess a vision-language model's capacity by analyzing the sources of errors in the visual-cognitive processes it employs when providing descriptions. A unique aspect of the presented work is that it could apply the technique to models spanning a decade, even though many of the early models cannot be easily made functional in current hardware. 
These errors might account for the actual gap in human and model visual reasoning capabilities in the pre-MLLMs and also help us probe humans and current MLLMs in scenes that require more complex reasoning.

\section*{Declaration of Generative AI and AI-assisted technologies in the writing process}
The author(s) wrote a draft of the paper.  The 1st author used Grammarly and ChatGPT in order to improve the readability and grammar throughout the manuscript. Author(s) further edited the content for clarity and tone. Author(s) take(s) take full responsibility for the content of the published article. 

\section*{Declaration of competing interests }
The authors declare no competing interests.

\clearpage
\appendix
\setcounter{figure}{0}
\renewcommand{\thefigure}{A.\arabic{figure}}
\section*{Supplementary}
\label{secA1}
\begin{figure}[!htbp]
    \centering
    \includegraphics[width=\textwidth, scale = 0.4]{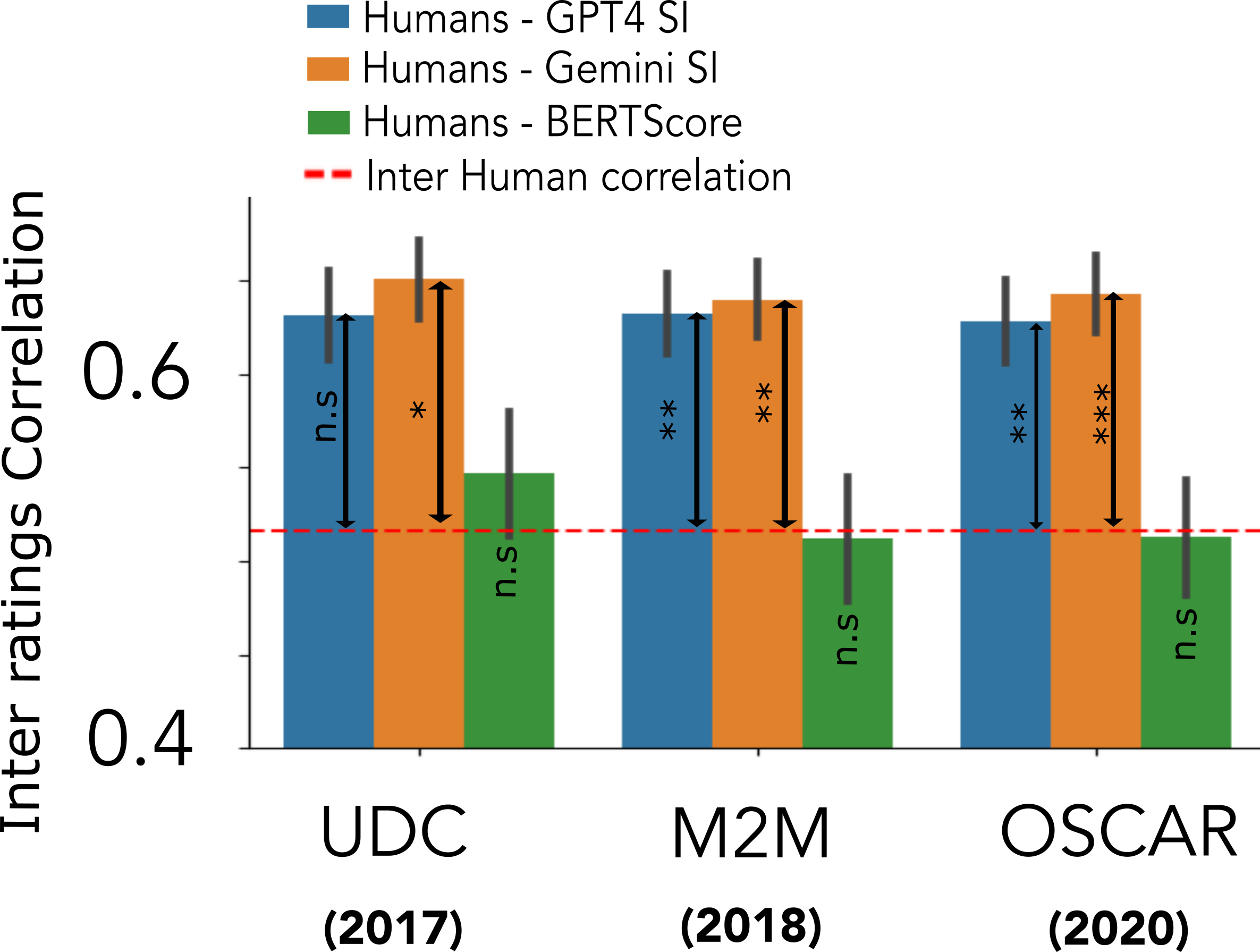}
    \caption{(a) Correlation of LLM-SI and BERTScore metrics with the human raters collected for comparing descriptions' similarity of UDC, M2M, and OSCAR models with the human gold standard description. The LLM-SI correlation was significantly higher than the average inter-human correlation among 20 human raters, suggesting that these metrics can serve as a substitute for human raters. The performance of BERTScore was on par with the inter-human correlation.}
    \label{fig:ExtData}
\end{figure}




\bibliographystyle{elsarticle-num}
\bibliography{bibliography}
\end{document}